\documentclass[11pt,a4paper]{article}
\usepackage{smile}
\usepackage[hyperref]{emnlp-ijcnlp-2019}
\usepackage{times}
\usepackage{latexsym}
\usepackage{xspace}
\usepackage{enumitem}

\usepackage{url}

\usepackage{booktabs} 
\usepackage{diagbox}
\usepackage{multirow}
\usepackage{balance}
\usepackage{amsfonts}
\usepackage{amsmath}
\usepackage{amssymb}
\usepackage{graphicx}
\usepackage{subfigure}
\usepackage{microtype}
\usepackage{xspace}
\usepackage{wrapfig,lipsum}
\usepackage{algorithm}
\usepackage{algorithmicx}
\usepackage{algpseudocode}
\usepackage{makecell}
\usepackage[T1]{fontenc}
\usepackage{soul}
\usepackage{dsfont}
\usepackage{tabularx, booktabs}
\newcolumntype{Y}{>{\centering\arraybackslash}X}
\newcolumntype{L}{>{\arraybackslash}X}

\aclfinalcopy 

\newcommand\DNAME{ALUM}
\newcommand{\robertabase}{RoBERTa$_{\textsc{base}}$}
\newcommand{\robertalarge}{RoBERTa$_{\textsc{large}}$}

\newcommand{\alumbert}{ALUM$_{\textsc{BERT-BASE}}$}
\newcommand{\alumrobertabase}{ALUM$_{\textsc{RoBERTa-BASE}}$}
\newcommand{\alumrobertalarge}{ALUM$_{\textsc{RoBERTa-LARGE}}$}

\newcommand{\eat}[1]{\ignorespaces}

\title{Adversarial Training for Large Neural Language Models}

\author{
Xiaodong Liu$^\dagger$, Hao Cheng$^\dagger$, Pengcheng He$^\ddagger$, Weizhu Chen$^\ddagger$, Yu Wang$^\dagger$, Hoifung Poon$^\dagger$, \\ \textbf{Jianfeng Gao$^\dagger$}
 \\ 
  $^\dagger$ Microsoft Research~~~~~~~~~~~~~~~~
  $^\ddagger$ Microsoft Dynamics 365 AI 
 \\
  {\tt \{xiaodl,chehao,penhe,wzchen,yuwan,hoifung,jfgao\}@microsoft.com}
}

\date{}

\begin{document}
\maketitle

\begin{abstract}

Generalization and robustness are both key desiderata for designing machine learning methods.
Adversarial training can enhance robustness, but past work often finds it hurts generalization.
In natural language processing (NLP), pre-training large neural language models such as BERT have demonstrated impressive gain in generalization for a variety of tasks, with further improvement from adversarial fine-tuning.
However, these models are still vulnerable to adversarial attacks.
In this paper, we show that adversarial pre-training can improve both generalization and robustness.
We propose a general algorithm {\DNAME} (\textbf{A}dversarial training for large neural \textbf{L}ang\textbf{U}age \textbf{M}odels), which regularizes the training objective by applying perturbations in the embedding space that maximizes the adversarial loss. 
We present the first comprehensive study of adversarial training in all stages, including pre-training from scratch, continual pre-training on a well-trained model, and task-specific fine-tuning. 
{\DNAME} obtains substantial gains over BERT on a wide range of NLP tasks, in both regular and adversarial scenarios. 
Even for models that have been well trained on extremely large text corpora, such as RoBERTa, {\DNAME} can still produce significant gains from continual pre-training, whereas conventional non-adversarial methods can not.
{\DNAME} can be further combined with task-specific fine-tuning to attain additional gains.
The {\DNAME} code is publicly available at \hyperlink{https://github.com/namisan/mt-dnn}{https://github.com/namisan/mt-dnn}.

\end{abstract}


\section{Introduction}
\label{sec:intro}

\eat{
The following three papers discuss the adversarial training and the tradeoff between robustness and generalization. 

https://arxiv.org/abs/2002.11080
https://arxiv.org/abs/2002.10716
https://arxiv.org/abs/1906.06032

The robust self-training (RST) https://arxiv.org/pdf/1905.13736.pdf is also shown to be able to improve robust accuracy without sacrificing standard accuracy for noiseless linear regression.

However, if we can show that ALUM improves over both BERT and RoBERTa, in both normal tasks and Adv-SQUAD and Hellaswag, then we have a pretty good storyline, as follows: 
-	Generalization and robustness are both core ML concerns
-	Adversarial training can enhance robustness, but past work often finds it hurts generalization
-	In NLP, much focus on brute-force way to run large-scale pretraining (scrape more data), but diminishing return
-	In this paper, we investigate adversarial pretraining, and found it improved BOTH generalization and robustness, for both standard and extremely large pretraining settings
-	It can also combines with adversarial fine-tuning for further gain

“in NLP,  previous works mostly applies adversarial training in fine-tuning, but this work firstly extends it to pre-training”

}

Generalization and robustness are two fundamental considerations in assessing machine learning methods.
Ideally, a learned model should perform well on unseen test examples and withstand adversarial attacks.
In natural language processing (NLP), pre-training neural language models on unlabeled text has proven very effective to improve generalization performance for a variety of downstream tasks, as exemplified by BERT \cite{devlin2018bert} and other transformer-based models \cite{liu2019roberta,gpt22019,clark2020electra,dong2019unified, bao2020unilmv2}. 
However, these models may still suffer catastrophic failures in adversarial scenarios \cite{nie2019adversarial, hsieh2019robust-self-att}. 
For example, \newcite{jin2019bertrobust} show that classification accuracy on a Yelp dataset drops from 95.6\% on standard test to 6.8\% on robust test for a BERT model. 

Adversarial training \cite{madry2017pgd,goodfellow2014explaining} has been well studied in computer vision, but past work shows that it often hurts generalization \cite{raghunathan2019adv-hurt,min2020curious}.
In NLP, there is growing interest in adversarial training, but existing work typically focuses on assessing the impact on generalization  \cite{zhu2019freelb,jiang2019smart,cheng-etal-2019-adv-nmt,wang2019adv-lm}. 
Moreover, adversarial training is generally limited to task-specific fine-tuning\footnote{A notable exception is \newcite{wang2019adv-lm}, but it only applied adversarial training to generative language modeling.}. See \newcite{minaee2020deepreview} for a recent survey.

In this paper, we present the first comprehensive study on adversarial pre-training, and show that it can improve both generalization and robustness for a wide range of NLP tasks.
We propose a unifying algorithm {\DNAME} (\textbf{A}dversarial training for large neural \textbf{L}ang\textbf{U}age \textbf{M}odels), which augments the standard training objective with an additional term that maximizes the adversarial loss via applying perturbation in the embedding space.
{\DNAME} is generally applicable 
to pre-training and fine-tuning, 
on top of any Transformer-based language models. 

We conduct a comprehensive evaluation on various NLP tasks across multiple benchmark datasets, including GLUE, SQuAD v1.1/v2.0, SNLI, SciTail for assessing model generalization, and ANLI, HELLSWAG, SWAG, Adversarial SQuAD for assessing model robustness.   
Experimental results show that by conducting adversarial pre-training, {\DNAME} attains significant improvements, often outperforming previous state of the art by a large margin.
This is true even for the extremely well-trained RoBERTa model, where continual pre-training without adversarial training fails to attain any gain.

Remarkably, in addition to improving generalization, we find that adversarial pre-training also substantially improves robustness, as exemplified by the resulting large gains in adversarial datasets such as ANLI, Adversarial-SQuAD, HELLASWAG, which significantly reduces the gap between standard errors and robust errors for popular models like BERT and RoBERTa. This suggests that adversarial training on unlabeled data can provide a promising direction to reconcile the apparent conflict between generalization and robustness as observed in prior work \cite{raghunathan2019adv-hurt,min2020curious}.
We also show that adversarial pre-training can be combined with adversarial fine-tuning, resulting in extra gains. 

Our contributions are summarized as follows:
\begin{itemize}
    \item We propose {\DNAME}, a general algorithm to incorporate adversarial training for pre-training and fine-tuning large neural language models.
    \item We conduct a comprehensive evaluation on a wide range of NLP tasks and assess the impact of adversarial training in pre-training from scratch, continual pre-training, task-specific fine-tuning, and their combinations.
    \item We obtain significant improvements over prior state of the art, including extremely well-trained models such as RoBERTa, in both generalization and robustness.
    \item To facilitate research, we will release our code and pre-trained models.
\end{itemize}

\section{Preliminary}
\label{sec:prelimiary}
In this section, we give a quick overview of language model pre-training, using BERT \cite{devlin2018bert} as a running example for transformer-based neural language models.

\subsection{Input Representation}
\label{subsec:IR}
We assume that the input consists of text spans (typically sentences) separated by a special token $[SEP]$. To address the problem of out-of-vocabulary words, tokens are divided into subword units, using Byte-Pair Encoding (BPE) \cite{sennrich2015bpe} or its variants \cite{kudo2018sentencepiece}, which generates a fixed-size subword vocabulary to compactly represent words in training text corpora.

\eat{
The model takes a sequence of sub-word tokens as inputs which contain two segments: $X_1=\{x_{11}, x_{12} ... x_{1m}\}$ and $X_2= \{x_{21}, x_{22} ... x_{2n}\}$ where n and m are the length of each segment respectively, sampled from a large-scale raw text corpus, e.g. Wikipedia.  In order to distinguish the boundary of these two segments, special tokens are added and it generates the final input sub-word tokens: $X=\{[CLS] x_{11}, x_{12} ... x_{1m} [SEP] x_{21}, x_{22} ... x_{2n}[SEP]\}$.
}

\subsection{Model Architecture}
\label{subsec:bert}
Following recent pre-training methods \cite{devlin2018bert,liu2019roberta}, 
we use transformer-based models \cite{vaswani2017attention} 
to leverage a multi-head attention mechanism, which have demonstrated superiority in parallel computation and modeling long-range dependencies, compared to recurrent neural networks such as LSTM \cite{hochreiter1997lstm}. 
The input is first passed to a lexical encoder, which combines 
a token embedding, a (token) position embedding and a segment embedding (i.e., which text span the token belongs to) by element-wise summation. 
The embedding layer is then passed to multiple layers of transformer modules to generate the contextual representation \cite{vaswani2017attention}. 

\subsection{Self Supervision}
\label{subsec:pre-train-task}
A key innovation in BERT \cite{devlin2018bert} is the use of \textbf{Masked Language Model (MLM)} for self-supervised pre-training.
Instead of predicting the next token based on the preceding tokens, as in traditional generative language models, MLM randomly replaces a subset of tokens by a special token (e.g., $[MASK]$), and asks the model to predict them. 
Essentially, it is a cloze task \cite{taylor1953cloze}, where the training objective is the cross-entropy loss between the original tokens and the predicted ones. In BERT and RoBERTa, 15\% of the input tokens are chosen, among which a random 80\% are replaced by $[MASK]$, 10\% are left unchanged and 10\% are randomly replaced by a token from the vocabulary. In our experiments, instead of using a fixed masked rate of 15\%, we gradually increase it from 5\% to 25\% with 5\% increment for every 20\% of training epochs, as we find this makes pre-training more stable.

\eat{
\noindent \textbf{Masked language modeling} is a cloze task \cite{taylor1953cloze}. During the training, we randomly select a bunch of tokens and replace them with a special token (e.g. [MASK]). The goal is that let the model to restore its original tokens which were replaced by [MASK]. Thus the training objective is a cross-entropy loss between the true masked tokens and predicting tokens. Originally, for example, BERT and RoBERTa select 15\% of the input tokens for possible replacement. Among them, 80\% of tokens are substituted by [MASK], 10\% are left unchanged and 10\% are replaced by a token which is randomly selected in vocabulary. In our experiments, we progressively increase the mask rate from 5\% to 25\% by 5\% in every 20\% of each training phrase, and find that it stabilizes training meanwhile keep the total masked tokens the same as the original training.
}

Additionally, BERT also uses \textbf{Next Sentence Prediction (NSP)}, which is a binary classification task that for a given sentence pair determines whether one sentence follows the other in the original text. There have debates on how much NSP helps \cite{liu2019roberta}. But we include it in our experiments for a fair comparison with BERT.
\section{{\DNAME} (\textbf{A}dversarial training for large neural \textbf{L}ang\textbf{U}age \textbf{M}odels)}
\label{sec:alum}

In this section, 
we first present a unifying view of standard training objectives and prior approaches to adversarial training. 
We then present {\DNAME}, which is a general adversarial training algorithm applicable to pre-training and fine-tuning, on top of any transformer-based neural language models.

\subsection{Standard Training Objectives}
\label{subsec:bg}
Both pre-training and fine-tuning can be viewed as minimizing the standard error on training data, with the training objectives derived from self-supervision (MLM and NSP in pre-training) or direct supervision (labeled examples in task-specific fine-tuning), respectively.

Specifically, the training algorithms seek to learn a function $f(x; \theta): x \rightarrow C$, parametrized by $\theta$. In MLM, $C$ is the vocabulary, and $f(x; \theta)$ tries to predict the masked token $y$. 
In fine-tuning, $C$ is the task-specific label set, and $f(x; \theta)$ is the classifier.
Given a training dataset $D$ of input-output pairs $(x,y)$ and the loss function $l(.,.)$ (e.g., cross entropy),
$f(x;\theta)$ is trained to minimize the empirical risk:
\begin{equation}
\min_{\theta} \mathbb{E}_{(x, y)\sim D}[l(f(x; \theta), y)]    
\label{eq:sl}
\end{equation} 

\subsection{Adversarial Training}
\label{subsec:at}
Pre-training a large neural language model such as BERT has proven effective to improve generalization performance in task-specific fine-tuning \cite{devlin2018bert}. However, such models can still suffer catastrophic loss in adversarial scenarios \cite{nie2019adversarial,hsieh2019robust-self-att,madry2017pgd,jin2019bertrobust}, with attacks as simple as replacing a few words in input sentences while preserving the semantics.

To improve model robustness and withstand adversarial attacks, adversarial training has been proposed and studied extensively, predominantly in computer vision literature \cite{goodfellow2014explaining,madry2017pgd}.
The key idea is to modify the training objective by applying small perturbation to input images that maximize the adversarial loss:
\begin{equation}
\min_{\theta} \mathbb{E}_{(x, y)\sim D}[\max_{
\delta} l(f(x + \delta; \theta), y)]
\label{eq:advt}
\end{equation}
where the inner maximization can be solved by running a number of projected gradient descent steps \cite{madry2017pgd}. 

While adversarial training has been successful in mitigating adversarial attacks, past work often encounters an apparent conflict between generalization and robustness \cite{raghunathan2019adv-hurt,raghunathan2020understanding,min2020curious}, as adversarial training could hurt generalization performance. 

\subsection{The {\DNAME} Algorithm}

In NLP, applying adversarial training is not straightforward, since the input are discrete elements (token or subword sequences), but there have been some recent successes  \cite{zhu2019freelb,jiang2019smart,cheng-etal-2019-adv-nmt,wang2019adv-lm,minaee2020deep}. 
However, aside from \newcite{wang2019adv-lm}, there has not been any prior work on adversarial pre-training, and \newcite{wang2019adv-lm} only applied adversarial training to generative language modeling using LSTM.

{\DNAME} is applicable to both pre-training and fine-tuning.
It builds on several key ideas that have proven useful in prior work.
First, instead of applying perturbation to the input text directly, one would perturb the embedding space. Namely, $x$ is the sub-word embedding in $f(x; \theta)$ \cite{jiang2019smart, zhu2019freelb}.

Second, instead of adopting the adversarial training objective of Eq.~\ref{eq:advt}, as in \newcite{zhu2019freelb} and most other approaches, we follow \newcite{jiang2019smart} to regularize the standard objective using virtual adversarial training \cite{miyato2018virtual}:
\begin{equation}
\begin{aligned}
  \min_{\theta} \mathbb{E}_{(x, y)\sim D}[l(f(x; \theta), y) + \\ 
  \alpha  \max_{
\delta} l(f(x+\delta; \theta), f(x; \theta))]
\end{aligned}
\label{eq:alum}
\end{equation}
Effectively, the adversarial term favors label smoothness in the embedding neighborhood, and $\alpha$ is a hyperparameter that controls the trade-off between standard errors and robust errors. 

We found that virtual adversarial training is superior to conventional adversarial training, especially when labels might be noisy. E.g., BERT pre-training uses the masked words as self-supervised labels, but in many cases, they could be replaced by other words to form completely legitimate text. Empirically, we verified that this is indeed the case, as pre-training benefits from larger $\alpha$. We set $\alpha=10$ for pre-training, and $\alpha=1$ for fine-tuning in all our experiments. 

Compared to standard training, adversarial training is rather expensive due to the inner maximization. \newcite{zhu2019freelb} adopted the {\em free adversarial training} idea in \newcite{shafahi2019freeat} for acceleration, by reusing the backward pass for gradient computation to carry out the inner ascent step and outer descent step simultaneously.
Inspired by ERNIE \cite{sun2019ernie} and other continual pre-training approaches, we instead adopt a curriculum learning approach: first train the model using the standard objective \eqref{eq:sl}; and then continue the training with virtual adversarial training \eqref{eq:alum}.

\newcite{jiang2019smart} also incorporated a momentum term using the Bregman proximate point method, which can be quite costly in training time. We found that our curriculum learning approach largely rendered this unnecessary  and simplified our algorithm without using this term.

\begin{algorithm}[!tb]
\caption{{\DNAME}}\label{algo:main}
\begin{algorithmic}[1]
	\INPUT $T$: the total number of iterations, $\cX=\{(x_1, y_1),...,(x_n, y_n)\}$:  the dataset,
	$f(x; \theta)$: the machine learning model parametrized by $\theta$, $\sigma^2$: the variance of the random initialization of perturbation $\delta$, $\epsilon$: perturbation bound, $K$: the number of iterations for perturbation estimation, $\eta$: the step size for updating perturbation, $\tau$: the global learning rate, $\alpha$: the smoothing proportion of adversarial training in the augmented learning objective, $\Pi$: the projection operation. 

	\For{$t=1,..,T$}
	\For{$(x,y) \in$ $\mathcal{X}$}
	    \State $\delta \sim \cN(0,\sigma^2I)$
    	\For{$m=1,..,K$}
    	    \State $g_{adv} \leftarrow \nabla_{\delta} l(f(x; \theta), f(x + \delta; \theta))$
    	    \State $\delta \leftarrow \Pi_{\|\delta\|_{\infty} \leq \epsilon} (\delta + \eta g_{adv})$  
    	\EndFor
	    \State $g_{\theta} \leftarrow \nabla_{\theta} l(f(x; \theta), y)$  
	    
	    $\qquad\quad+ \alpha \nabla_{\theta}l(f(x; \theta), f(x + \delta; \theta))$ 
	    \State $\theta \leftarrow \theta - \tau g_{\theta}$
	\EndFor
	\EndFor
		\OUTPUT $\theta$
\end{algorithmic}
\end{algorithm}

Algorithm~\ref{algo:main} shows the details of {\DNAME}. Line 4-6 run $K$ projected gradient steps to find the perturbation $\delta$ that maximizes the adversarial loss (violation of local smoothness). Note that a larger $K$ leads to better approximation \cite{madry2017pgd,qin2019adversarial}, but it is more expensive. To attain a good trade-off between speed and performance, we set $K=1$ in all our experiments. 

\subsection{Generalization vs. Robustness}
Empirically, we found that by applying adversarial pre-training using {\DNAME}, we were able to improve both generalization and robustness for a wide range of NLP tasks, as seen in Section~\ref{sec:exp}. 
This is very interesting as prior work often finds that adversarial training hurts generalization, even with theoretical justification \cite{raghunathan2019adv-hurt,raghunathan2020understanding,min2020curious}. 

We hypothesize that adversarial pre-training might be the key for reconciling this apparent incongruence, as prior work on the conflict between generalization and robustness generally focuses on the supervised learning setting. Interestingly, some nascent results in reconciling the two also leverage unlabeled data, such as self-training \cite{raghunathan2020understanding}.
Additionally, we hypothesize that by perturbing the embedding space rather than the input space, adversarial training in NLP might inadvertently bias toward on-manifold perturbation than regular perturbation, which helps generalization \cite{stutz2019disentangling}. 
 We leave the theoretical analysis of all these connections to future work.


\section{Experiments}
\label{sec:exp}
\vspace{-2mm}
In this section, we present a comprehensive study of adversarial training on large neural language models. We show that {\DNAME} substantially improves both generalization and robustness in a wide range of NLP tasks, for both the standard BERT model and the extremely well-trained RoBERTa model. We also show that {\DNAME} can be applied to adversarial pre-training and fine-tuning alike and attain further gain by combining the two.

\subsection{Datasets}
\label{subsec:dataset}
\paragraph{Pre-training:} For BERT pre-training, we use Wikipedia (English Wikipedia dump\footnote{https://dumps.wikimedia.org/enwiki/}; 13GB). For continual pre-training of RoBERTa, we use Wikipedia (13GB), OPENWEBTEXT (public Reddit content \cite{Gokaslan2019OpenWeb}; 38GB), STORIES (a subset of CommonCrawl \cite{trinh2018simple}; 31GB).

\paragraph{NLP application benchmarks:} 
To assess the impact of adversarial training on generalization, we use standard benchmarks such as GLUE \cite{wang2018glue} and SQuAD (v1.1 and v2.0) \cite{squad1,squad2}, as well as three named entity recognition (NER) tasks in the biomedical domain.
To evaluate the impact of adversarial training on robustness, we use ANLI~\cite{nie2019adversarial}, 
Adversarial SQuAD~\cite{jia2017advsquad}, and 
HELLASWAG~\cite{hampel1974influence}. 
To assess the combination of adversarial pre-training and fine-tuning, we follow \newcite{jiang2019smart} and use MNLI~\cite{mnli2018} (from GLUE), ANLI, SWAG~\cite{zellers2018swag}, SNLI~\cite{snli2015}, SciTail~\cite{scitail}. 
These benchmarks cover a wide range of NLP tasks such as named entity recognition, textual entailment, and machine reading comprehension, spanning classification, ranking, and regression. For details, see Appendix~\ref{sec:appendix}.

\subsection{Implementation Details}
\label{subsec:details}
We perform three types of adversarial training in our experiments: pre-training from scratch, continual pre-training on a well-trained model, and task-specific fine-tuning.

We pre-train BERT models from scratch using Wikipedia\footnote{BookCorpus is no longer publicly available.}. 
The training code is based on Megatron, implemented in PyTorch  \cite{shoeybi2019megatron}\footnote{https://github.com/NVIDIA/Megatron-LM}.  
We use ADAM \cite{kingma2014adam} for the optimizer with a standard learning rate schedule that increases linearly from zero to the peak rate of $1\times 10^{-4}$ in first one percent of steps, and then decays linearly to zero in the remaining 99\% of steps. 
Following \citet{devlin2018bert}, training is done for one million steps with batch size of 256.
We set the perturbation size $\epsilon=1\times 10^{-5}$, the step size $\eta=1\times10^{-3}$, and the variance for initializing perturbation $\sigma=1\times10^{-5}$. 
We set $\alpha=10$ for heightened regularization in virtual adversarial training, and set $K=1$ for training efficiency (i.e., one projected gradient step). 
The training takes 10 days on one DGX-2 machine with 16 V100 GPUs.

For continual pre-training of RoBERTa \cite{liu2019roberta}, 
we use RoBERTa's default training parameters, except a smaller learning rate ($4\times10^{-5}$), and run for 100K training steps with a batch size of 256 on the union of Wikipedia, OPENWEBTEXT, and STORIES (total size 82GB). 
The code is based on FairSeq\footnote{https://github.com/pytorch/fairseq}. 
The training takes 7 days on two DGX-2 machines.

For fine-tuning with or without adversarial training,
we use the MT-DNN open-sourced toolkit \cite{liu2020mtmtdnn,liu2015mtl}\footnote{https://github.com/namisan/mt-dnn}. 
We follow \citet{jiang2019smart} for head-to-head comparison, which uses ADAM \cite{kingma2014adam} and RADAM \cite{liu2019radam} as our optimizers, with peak learning rates of $\{5\times 10^{-6}, 8 \times 10^{-6}, 1\times 10^{-5}, 2\times 10^{-5}\}$, and batch sizes of {16, 32 or 64}, depending on the tasks.  
The dropout rate is set to $0.1$ for all the task-specific layers, except $0.3$ for MNLI and $0.05$ for CoLA. 
To avoid gradient exploding, the gradient is clipped to keep the norm within $1$.  
All the texts are tokenized using WordPiece and chopped to spans up to $512$ tokens. 
We conduct fine-tuning for up to 10 epochs and pick the best model using the dev set.

\subsection{Improving Generalization}
\label{subsec:gen}

In this subsection, we study the impact of adversarial pre-training on generalization, by comparing the performance of pre-trained models in various downstream applications. First, we study the scenario of pre-training from scratch, by comparing three BERT models:
\begin{itemize}
    \item \textbf{BERT\textsubscript{BASE}} is the standard BERT base model trained using the same setting as \citet{devlin2018bert} (i.e., 1M steps with a batch size of 256).
    \item \textbf{BERT+\textsubscript{BASE}} is similar to {BERT\textsubscript{BASE}}, except that it is trained with 1.6M steps, which takes roughly the same amount of time as that of adversarial pre-training (see {\alumbert{}} below). 

    \item \textbf{\alumbert{}} is a BERT model trained using the same setting as {BERT\textsubscript{BASE}}, except that {\DNAME} is used in the last 500K steps. Each adversarial training step takes approximately 1.5 times longer than a step in standard training\footnote{With K=1 in Algorithm~\ref{algo:main}, {\DNAME} requires two more forward passes and one more backward pass compared to standard training.}.
\end{itemize}
\begin{table}[htb!]
\begin{tabular}{@{\hskip3pt}l@{\hskip3pt}|@{\hskip3pt} c@{\hskip3pt} @{\hskip3pt}c@{\hskip3pt}|@{\hskip3pt}c@{\hskip3pt}}
\toprule
\multirow{3}{*}{\bf Model} & \multicolumn{2}{@{\hskip2pt}l|@{\hskip3pt}}{SQuAD v1.1/v2.0} &{MNLI} \\ 
& &  & { m/mm} \\ \cline{2-4}
& F1/EM & F1/EM & {Acc} \\

\midrule
BERT\textsubscript{BASE} & 88.5/81.0 &76.5/72.9  & 84.5/84.4 \\ \hline
BERT+\textsubscript{BASE} & 89.6/82.4 & 77.8/74.0  & 85.0/84.8 \\ \hline
\alumbert{} &\textbf{90.8/83.7} & \textbf{80.2/76.6} & \textbf{85.8/86.1}  \\
\bottomrule
\end{tabular}
\caption{
Comparison of standard and adversarial pre-training on SQuAD (v1.1 and v2.0) and MNLI (in-domain and out-domain). BERT\textsubscript{BASE} and \alumbert{} both use 1M pre-training steps, and BERT+\textsubscript{BASE} use 1.6M steps.
}
\label{tab:main}
\end{table}

\begin{figure}[ht]
	\centering
    {
	\includegraphics[width=0.49\textwidth]{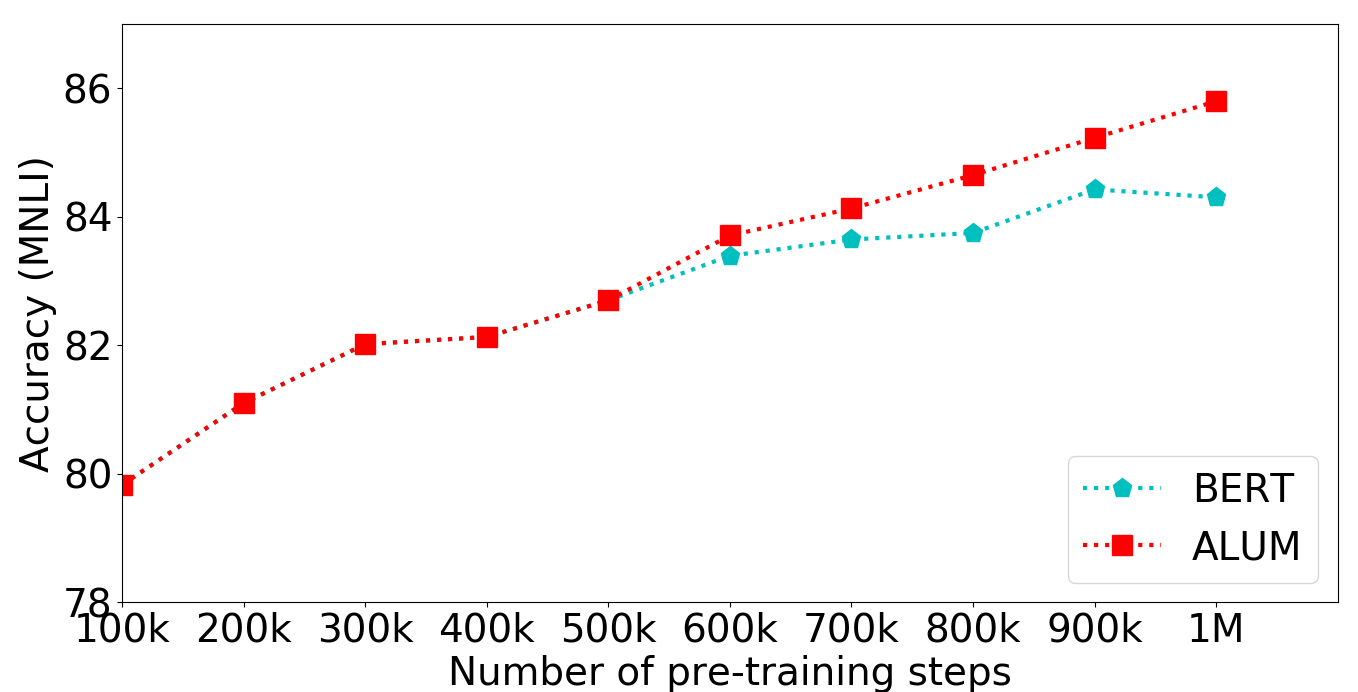}
    }
	\caption{Comparison of the standard and adversarial pre-training on the MNLI development set.}
	\label{fig:alum}
\end{figure}
\begin{table}[ht]
    \centering
    \begin{tabular}{@{\hskip2pt}l@{\hskip3pt}|@{\hskip3pt}l@{\hskip3pt}|@{\hskip3pt}c@{\hskip3pt}|@{\hskip3pt}c@{\hskip2pt}}
    \toprule
    Model & BC2GM & NCBI & JNLPBA \\
          & F1 & F1 & F1 \\ 
          \hline
    BERT\textsubscript{BASE} &79.7 &84.6& 75.7 \\ \hline      
    \alumbert{} &\textbf{81.1} & \textbf{86.9} &  \textbf{76.5} \\
        \bottomrule

    \end{tabular}
    \caption{Comparison of standard and adversarial pre-training on biomedical NER. Scores are entity-level F1.}
    \label{tab:biomed_eval}
\end{table}

Table~\ref{tab:main} compares these pre-trained models on three standard benchmarks (SQuAD v1.1 \cite{squad1} and v2.0 \cite{squad2}, and MNLI from GLUE \cite{wang2018glue}), using the same standard fine-tuning setting (without adversarial training).
The standard BERT models trained using only the Wikipedia data attain similar results as in \citet{devlin2018bert}, thus provide a good baseline for comparison.
\alumbert{} consistently outperforms the standard BERT models across all the datasets, even adjusting for the slightly longer trainng time. E.g., on SQuAD v1.1, \alumbert{} gains 2.3\% points in F1 over {BERT\textsubscript{BASE}} and 1.2\% points over {BERT+\textsubscript{BASE}}.
Figure~\ref{fig:alum} shows {\DNAME} at work on the development set of MNLI. Once adversarial training is applied in the middle (after first 500K steps), {\DNAME} starts outperforming BERT and the gap is widening.

We also assess the impact of adversarial pre-training in the biomedical domain, which is substantially different from the Wikipedia corpus used in pre-training.
Table~\ref{tab:biomed_eval} shows the results on standard biomedical name entity recognition (NER) datasets: BC2GM \cite{smith2008bc2gm}, NCBI \cite{dogan2014ncbi}, JNLPBA \cite{collier-kim-2004-JNLPBA}. Interestingly, {\DNAME} still outperforms the standard BERT model on all three tasks, even though the application domain is substantially different from the pre-training one.


\begin{table}[t]
\centering
\begin{tabular}{l|c|c}
\toprule
\multirow{2}{*}
{\bf Model} & MNLI-m/mm &SST-2 \\ 
& Acc & Acc\\
\midrule
\robertalarge{}& 90.2/90.2 & 96.4\\ \hline
RoBERTa+\textsubscript{LARGE} & 90.3/90.2  & 96.3\\
\bottomrule
\end{tabular}
\caption{
RoBERTa is an extremlly well-trained model: standard continual pre-training without adversarial training fails to improve generalization performance in downstream tasks. (Scores are accuracy.)
}
\label{tab:roberta_ct}
\end{table}

\begin{table*}[t]
\begin{center}
\begin{tabular}{@{\hskip1pt}l@{\hskip1pt}|@{\hskip1pt}c|@{\hskip1pt}c|@{\hskip1pt}c|@{\hskip1pt}c|@{\hskip1pt}c|@{\hskip1pt}c|@{\hskip1pt}c|@{\hskip1pt}c}
\toprule
\textbf{Model}         & \textbf{MNLI-m} & \textbf{SST-2} & \textbf{QNLI} & \textbf{CoLA} & \textbf{RTE}  & \textbf{MRPC} & \textbf{QQP}  & \textbf{STS-B} \\ \midrule
\robertabase{}     &      87.6       &      94.8      &     92.8      & \textbf{63.6} & 78.7 &     90.2      & 91.9 & \textbf{91.2} \\ 
\alumrobertabase{}     &\textbf{88.1}    &\textbf{95.3}           &\textbf{93.1}          & \textbf{63.6} &\textbf{80.2} &      \textbf{90.9}& \textbf{92.0} &  91.1\\ \hline
\robertalarge{}        &      90.2       &      96.4      &  94.7        & 67.8 & 86.6 &     90.9      & \textbf{92.2} & \textbf{92.4} \\ 
{\alumrobertalarge{}}     &\textbf{90.9}    & \textbf{96.6}         &  \textbf{95.1}         & \textbf{68.2} & \textbf{87.3}&      \textbf{91.1}& \textbf{92.2} &  92.1\\
\bottomrule
\end{tabular}
\end{center}
\caption{
Comparison of standard and adversarial pre-training on the GLUE development set.
Results for {\alumrobertabase{}} and {\alumrobertalarge{}} are averaged over five runs. Results of \robertabase{} and \robertalarge{} are taken from \citet{liu2019roberta}.
}
\label{tab:glue:dev}
\end{table*}

\begin{table*}[htb!]
    \centering
    \begin{tabular}{l|c|c|c|c|c|c|c|c}
    \hline
\toprule    
         \multirow{2}{*}{Method} & \multicolumn{4}{c|}{Dev} & \multicolumn{4}{c}{Test}  \\
         \cline{2-9}
          &  R1 & R2 & R3 & All & R1 & R2 & R3 & All  \\
          \hline 
 		\multicolumn{9}{c}{ MNLI + SNLI + ANLI + FEVER  }  \\ \hline
 BERT\textsubscript{BASE}&55.7&46.3&43.4 &48.2 &55.1 &45.0 &43.1 & 47.4\\ \hline
 BERT+\textsubscript{BASE}&57.5 &47.3 & 43.0& 48.9&57.7 &43.7 & 43.0&47.8\\ \hline
\alumbert{} &62.0&48.6&48.1 & \textbf{52.6} &61.3 &45.9 &44.3 & \textbf{50.1}\\ \hline \hline			
          BERT\textsubscript{LARGE}  \citep{nie2019adversarial} &57.4&48.3&43.5& 49.3 &-&-&-& 44.2 \\
		\hline
		XLNet\textsubscript{LARGE} \citep{nie2019adversarial}&67.6&50.7&48.3& 55.1 &-&-&-& 52.0 \\
		\hline
		RoBERTa\textsubscript{LARGE} \citep{nie2019adversarial} &73.8&48.9&44.4& 53.7 &-&-&-& 49.7 \\
		\hline 
	    \alumrobertalarge{} & 73.3&53.4&48.2& \textbf{57.7} &72.3&52.1&48.4& \textbf{57.0} \\
\bottomrule
    \end{tabular}
    \caption{Comparison of standard and adversarial pre-training on the adversarial dataset ANLI. R1, R2 and R3 are rounds with increasing difficulty. Note that \citet{nie2019adversarial} did not represent results for individual rounds, as signified by ``-''.}
    \label{tab:anli_full}
\end{table*}

\begin{table*}[htb!]
    \centering
    \begin{tabular}{l|c|c||c|c}
    \hline
\toprule    
         \multirow{3}{*}{Method} & 
         \multicolumn{2}{c||}{Adversarial SQuAD} & \multicolumn{2}{c}{HELLASWAG}  \\
         \cline{2-5}
          & AddSent & AddOneSent  &  Dev & Test \\ \cline{2-5}
          &EM/F1&EM/F1 & Accuracy &Accuracy\\ \hline
          BERT\textsubscript{BASE} &  48.9/54.0	&59.0/64.8 & 39.5 & -\\
          BERT+\textsubscript{BASE} & 50.1/56.2 	&60.5/65.7 & 40.3 & -\\
          \alumbert{} &  \textbf{54.6/60.4} &	\textbf{63.2/69.8} &\textbf{44.0} & -\\ \hline \hline
           RoBERTa\textsubscript{LARGE} &	72.3/66.0&79.3/72.9 & 85.0 &85.2 \\
          \alumrobertalarge{}&\textbf{75.5/69.4}&	\textbf{81.4/75.0}	&\textbf{86.2}& \textbf{85.6}\\ 
\bottomrule
    \end{tabular}
    \caption{Comparison of standard and adversarial pre-training on adversarial datasets Adversarial SQuAD and HELLASWAG. The test result on HELLASWAG is taken from the official leaderboard: \hyperlink{ rowanzellers.com/hellaswag}{rowanzellers.com/hellaswag}; we couldn't get results for BERT base models as the organizers restrict the number of submissions.}
    \label{tab:squad}    
\end{table*}

Next, we assess the impact of adversarial training in the continual pre-training setting. 
We use our pre-training dataset (Wikipedia, OPENWEBTEXT, STORIES; 82GB)\footnote{This is a subset of the data (160GB) used in RoBERTa pre-training.}, and run 100K steps in all our continual pre-training experiments.
We choose the RoBERTa models as the baseline, which use the same neural model as BERT, but were pre-trained on an order of magnitude more text (160GB vs 13GB). They are the state-of-the-art pre-trained language models, outperforming the standard BERT models in many NLP tasks.

RoBERTa models are extremely well-trained. Standard continual pre-training fails to attain any gains in downstream applications such as MNLI \cite{mnli2018} and SST \cite{sst2013} from GLUE \cite{wang2018glue}, as shown in Table~\ref{tab:roberta_ct}. 
On the other hand, {\DNAME} is able to attain further gain from continual pre-training of RoBERTa, as shown in Table~\ref{tab:glue:dev}. E.g., {\alumrobertabase{}} outperforms RoBERTa\textsubscript{BASE} by +0.5\%, and {{\alumrobertalarge{}}} outperforms \robertalarge{} by +0.7\% on the MNLI development set. This is rather remarkable, as by contrast standard continual pre-training is unable to attain any gain.

\subsection{Improving Robustness}
\label{subsec:ct}
In this subsection, we assess the impact of adversarial pre-training on the model's robustness against adversarial attacks, using three standard adversarial NLP benchmarks: ANLI \cite{nie2019adversarial}, HELLASWAG \cite{zellers2019hellaswag} and adversarial SQuAD \cite{jia2017advsquad}. 
On ANLI, we follow the experimental setting of \citet{nie2019adversarial} to enable a head-to-head comparison, which combines four datasets (ANLI, MNLI, SNLI and FEVER~\cite{thorne2018fever}) for fine-tuning.

Adversarial pre-training substantially improves model robustness, as shown in Table~\ref{tab:anli_full} and Table~\ref{tab:squad}.
In all three adversarial datasets, {\DNAME} consistently outperformed the standard pre-training counterparts, for BERT and RoBERTa alike.
For example, on ANLI, {\alumrobertalarge{}} gains 7.3\% points in test accuracy over \robertalarge{}, outperforms XLNet~\cite{yang2019xlnet} by 5.0\% points, creating a new state-of-the-art result.
The gains on Adversarial SQuAD and HELLASWAG are equally significant.
For example, for Adversarial SQuAD, \alumbert{} outperforms BERT\textsubscript{BASE} by +6.4\% F1 in the AddSent setting and +5.0\% F1 in the AddOneSent setting. 
Against \robertalarge{}, \alumrobertalarge{} gains +3.4\% F1 in AddSent and +2.1\% F1 in AddOneSent. 

\subsection{Combining Adversarial Pre-Training and Fine-tuning}
\label{subsec:aft}
\begin{figure}[htb!]
\centering  
\subfigure[Results on MNLI]{
\includegraphics[width=0.99\linewidth]{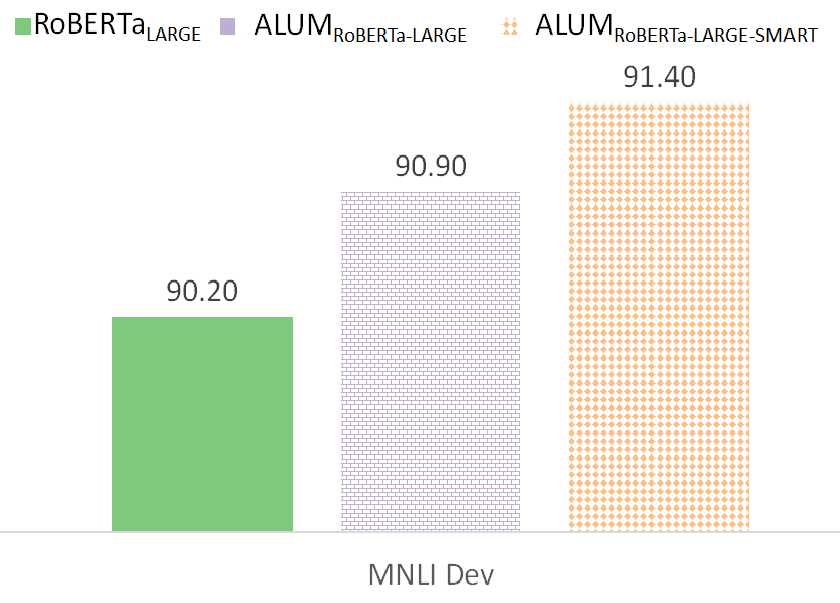}
}
\subfigure[Results on ANLI]{\includegraphics[width=0.98\linewidth]{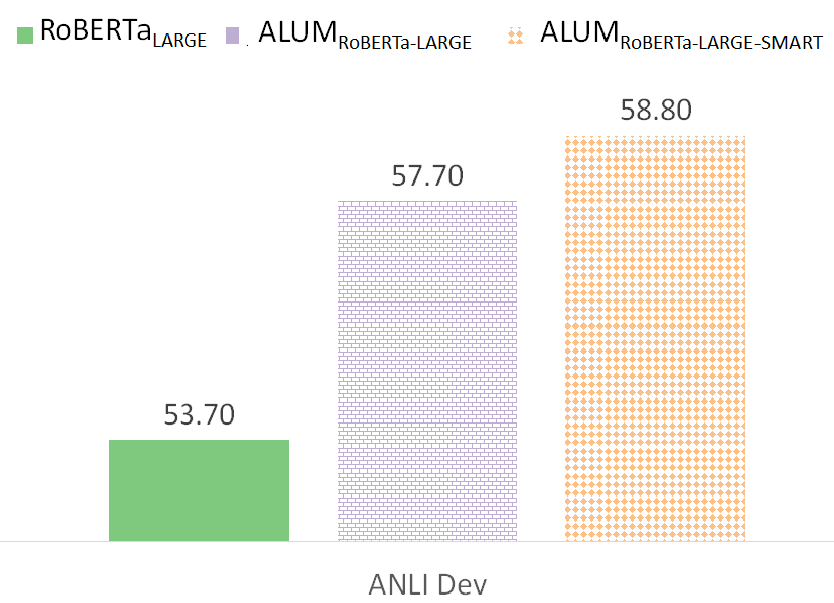}}
	\caption{
Combining adversarial pre-training and fine-tuning attaining the best results on the development sets of MNLI and ANLI, two representative GLUE tasks.
	}
	\label{fig:adv-ft}	
\end{figure}

\begin{table}[htb!]
	\begin{center}
		\begin{tabular}{@{\hskip1pt}l | c | c @{\hskip1pt}}
	\toprule    
	
	   \bf Model &Dev& Test  \\ \hline 
		\multicolumn{3}{c}{ SNLI Dataset (Accuracy\%)}  \\ \hline 
		GPT \cite{gpt22019} &- & 89.9 \\ \hline
		BERT\textsubscript{LARGE} &91.7& 91.0\\ \hline		
		MT-DNN\textsubscript{LARGE}\cite{liu2019mt-dnn}&92.2& 91.6\\ \hline		
		{\alumrobertalarge{}} &93.1& 93.0\\ \hline
		{\alumrobertalarge{}}\textsubscript{-SMART} &\textbf{93.6}& \textbf{93.4}\\ \hline
		\hline		
		\multicolumn{3}{c}{ SciTail Dataset (Accuracy\%)}  \\ \hline
		GPT \cite{gpt22019} &- &88.3 \\ \hline
		BERT\textsubscript{LARGE}\cite{liu2019mt-dnn} &95.7& 94.4\\ \hline		
		MT-DNN\textsubscript{LARGE}\cite{liu2019mt-dnn} &96.3& 95.0\\ \hline
		{\alumrobertalarge{}} &97.4 & 96.3\\ \hline
		{\alumrobertalarge{}}\textsubscript{-SMART} & \textbf{98.2}&\textbf{96.8} \\
		\bottomrule
		\end{tabular}
	\end{center}
	\caption{Combining adversarial pre-training and fine-tuning attains the best results on SNLI and SciTail. 
	}
	\label{tab:snli}
\end{table}

\begin{table}[htb!]
	\begin{center}
		\begin{tabular}{@{\hskip1pt}l | c | c @{\hskip1pt}}
	\toprule    
	
	   \bf Model &Dev& Test  \\ \hline 

		\multicolumn{3}{c}{ SWAG Dataset (Accuracy\%)}  \\ \hline 
		GPT \cite{gpt22019} &- & 78.0 \\
		\hline
		BERT\textsubscript{LARGE} \cite{devlin2018bert} &-& 86.3\\ \hline		
		Human\cite{zellers2018swag} & 88.0 &88.0 \\ \hline
		RoBERTa\textsubscript{LARGE} \cite{liu2019roberta} &-& 89.9\\ \hline				
		{\alumrobertalarge{}}&90.7& \textbf{91.0}\\ \hline	
		{\alumrobertalarge{}}\textsubscript{-SMART}&\textbf{91.2}& -\\ \hline		
		\hline
		\multicolumn{3}{c}{HELLASWAG Dataset (Accuracy\%)}  \\ \hline 
		GPT \cite{zellers2019hellaswag} &41.9 &41.7 \\ \hline
		BERT\textsubscript{LARGE}\cite{zellers2019hellaswag} &46.7& 47.3\\ \hline		
RoBERTa\textsubscript{LARGE}\cite{liu2019roberta} &-& 85.2\\ \hline		
		{\alumrobertalarge{}} &86.2& \textbf{85.6}\\ \hline
		{\alumrobertalarge{}}\textsubscript{-SMART}&\textbf{86.9}& -\\ \hline \hline	
		Human & 95.7 &95.6 \\ 	
		\bottomrule
		\end{tabular}
	\end{center}
	\caption{Combining adversarial pre-training and fine-tuning attains the best results on SWAG and HELLASWAG.}
	\label{tab:swag}
\end{table}

Adversarial training has been shown to be effective in task-specific fine-tuning \cite{jiang2019smart, zhu2019freelb}. In this subsection, we explore combining adversarial pre-training with adversarial fine-tuning.  
Specifically, we use \robertalarge{} as the base model, and compare it with \alumrobertalarge{}, which uses adversarial continual pre-training but standard fine-tuning, and 
{\DNAME}\textsubscript{RoBERTA-LARGE-SMART}, which uses adversarial training in both continual pre-training and fine-tuning. 
Figure~\ref{fig:adv-ft} shows the results on the development sets of MNLI and ANLI, two representative GLUE tasks. 
Combining adversarial pre-training and fine-tuning attains the best results, and substantially outperforms \robertalarge{}.
E.g., on ANLI, {\DNAME\textsubscript{RoBERTa-SMART}} outperforms \alumrobertalarge{} by +1.1\% points in accuracy, and outperforms \robertalarge{} by +5.1\% points.
On SNLI, SciTail, SWAG, and HELLASWAG, we observe similar gains by combining adversarial pre-training and fine-tuning, attaining new state-of-the-art results on these tasks. See table~\ref{tab:snli} and \ref{tab:swag}.


\section{Conclusion}
\label{sec:conclusion}
We propose {\DNAME}, a general adversarial training algorithm, and present the first comprehensive study of adversarial training in large neural language models. We show that adversarial pre-training can significantly improves both generalization and robustness, which provides a promising direction for reconciling their conflicts as observed in prior work. {\DNAME} substantially improved accuracy for BERT and RoBERTa in a wide range of NLP tasks, and can be combined with adversarial fine-tuning for further gain.

Future directions include: further study on the role of adversarial pre-training in improving generalization and robustness; speed up adversarial training; apply {\DNAME} to other domains.

\eat{
In this study, we explore adversarial training in the realm of large-scale neural language model pre-training and have shown that the proposed {\DNAME} algorithm improves both model generalization and robustness for a wide range of NLP tasks, and  
that combing {\DNAME} with task-specific adversarial fine-tuning creates new SOTAs for a number of NLP tasks, such as SWAG and SNLI.
One drawback of {\DNAME} is its computational cost, which makes pre-training even more time-consuming. 
In future work, we will explore methods to speed up the adversarial pre-training.
}
\section*{Acknowledgments}
We thank Haoming Jiang, Tuo Zhao, Zhe Gan, Keivn Duh, Yangfeng Ji, Greg Yang, Pengchuan Zhang, Lei Zhang, Furu Wei, Li Dong, Masayuki Asahara, and Lis Pereira for valuable discussions and comments, Microsoft Research Technology Engineering team for setting up GPU machines. 

\bibliography{ref}
\bibliographystyle{acl_natbib}
\clearpage
\appendix
\section{NLP Application Benchmarks}
\label{sec:appendix}
\begin{table*}[!htb]
	\begin{center}
		\begin{tabular}{l|l|c|c|c|c|c}
			\toprule 
			\bf Corpus &Task& \#Train & \#Dev & \#Test   & \#Label &Metrics\\ \hline \hline
			\multicolumn{6}{@{\hskip1pt}r@{\hskip1pt}}{Single-Sentence Classification (GLUE)} \\ \hline
			CoLA & Acceptability&8.5k & 1k & 1k & 2 & Matthews corr\\ \hline
			SST & Sentiment&67k & 872 & 1.8k & 2 & Accuracy\\ \hline \hline
			\multicolumn{6}{@{\hskip1pt}r@{\hskip1pt}}{Pairwise Text Classification (GLUE)} \\ \hline
			MNLI & NLI& 393k& 20k & 20k& 3 & Accuracy\\ \hline
            RTE & NLI &2.5k & 276 & 3k & 2 & Accuracy \\ \hline
            WNLI & NLI &634& 71& 146& 2 & Accuracy \\ \hline
			QQP & Paraphrase&364k & 40k & 391k& 2 & Accuracy/F1\\ \hline
            MRPC & Paraphrase &3.7k & 408 & 1.7k& 2&Accuracy/F1\\ \hline
			QNLI & QA/NLI& 108k &5.7k&5.7k&2& Accuracy\\ \hline \hline
			\multicolumn{5}{@{\hskip1pt}r@{\hskip1pt}}{Text Similarity (GLUE)} \\ \hline
			STS-B & Similarity &7k &1.5k& 1.4k &1 & Pearson/Spearman corr\\ \hline
			\multicolumn{6}{@{\hskip1pt}r@{\hskip1pt}}{Pairwise Text Classification} \\ \hline
			SNLI & NLI& 549k &9.8k&9.8k&3& Accuracy\\ \hline
			SciTail & NLI& 23.5k &1.3k&2.1k&2& Accuracy\\ \hline
			ANLI & NLI& 163k &3.2k&3.2k&3& Accuracy\\ \hline \hline
			\multicolumn{5}{@{\hskip1pt}r@{\hskip1pt}}{Span Classification} \\ \hline
			SQuAD v1.1 & MRC&  87.6k& 10.5k&9.5k &-& Exact Match (EM)/F1\\ \hline 
			SQuAD v2.0 & MRC& 130.3k &11.9k& 8.9k&-& Exact Match (EM)/F1\\ \hline 
			\multicolumn{5}{@{\hskip1pt}r@{\hskip1pt}}{Ranking} \\ \hline
			SWAG & Multiple choice&  73.5k& 20k&20k &-& Accuracy\\ \hline 
			HELLASWAG & Multiple choice&34k  &10& 10k&-& Accuracy\\ \hline
			\multicolumn{5}{@{\hskip1pt}r@{\hskip1pt}}{Biomedical Domain} \\ \hline
			BC2GM & NER &  12.6k& 2.5k & 5k &-& F1/Precision/Recall\\ \hline 
			NCBI & NER &  5.4k & 923 & 940 &-& F1/Precision/Recall\\ \hline 
			JNLPBA & NER &  14.7k & 3.9k & 3.9k &-& F1/Precision/Recall\\ 
			\bottomrule

		\end{tabular}
	\end{center}
	\caption{Summary information of the NLP application benchmarks.
	}
	\label{tab:datasets}
\end{table*}
\noindent $\bullet$ \textbf{GLUE}. The General Language Understanding Evaluation (GLUE) benchmark is a collection of nine natural language understanding (NLU) tasks. As shown in Table~\ref{tab:datasets},
it includes question answering~\cite{squad1}, linguistic acceptability~\cite{cola2018}, sentiment analysis~\cite{sst2013}, text similarity~\cite{sts-b2017}, paraphrase detection~\cite{mrpc2005}, and natural language inference (NLI)~\cite{rte1,rte2,rte3,rte5,winograd2012,mnli2018}. The diversity of the tasks makes GLUE very suitable for evaluating the generalization and robustness of NLU models. 

\noindent $\bullet$ \textbf{SNLI}.
The Stanford Natural Language Inference (SNLI) dataset contains 570k human annotated sentence pairs, in which the premises are drawn from the captions of the Flickr30 corpus and hypotheses are manually annotated \cite{snli2015}. 
This is the most widely used entailment dataset for NLI.

\noindent $\bullet$ \textbf{SciTail}.
This is a textual entailment dataset derived from a science question answering (SciQ) dataset \cite{scitail}. The task involves assessing whether a given premise entails a given hypothesis.  
In contrast to other entailment datasets mentioned previously, the hypotheses in SciTail are created from science questions while the corresponding answer candidates and premises come from relevant web sentences retrieved from a large corpus. As a result, these sentences are linguistically challenging and the lexical similarity of premise and hypothesis is often high, thus making SciTail particularly difficult. 

\noindent $\bullet$ \textbf{ANLI}.
The Adversarial Natural Language Inference (ANLI, \citet{nie2019adversarial}) is a new large-scale NLI benchmark dataset, collected via an iterative, adversarial human-and-model-in-the-loop procedure. Specifically, the instances are chosen to be difficult for the state-of-the-art models such as BERT and RoBERTa.

\noindent $\bullet$ \textbf{SWAG}. It is a large-scale adversarial dataset for the task of grounded commonsense inference, which unifies natural language inference and physically grounded reasoning \cite{zellers2018swag}. SWAG consists of 113k multiple choice questions about grounded situations. 

\noindent $\bullet$ \textbf{HELLASWAG}. It is similar to SWAG but more challenging \cite{zellers2019hellaswag}. For each query in HELLASWAG, it also has 4 choices and the goal is to find the best choice among them.

\noindent $\bullet$ \textbf{SQuAD v1.1/v2.0}. Stanford Question
Answering Dataset (SQuAD) v1.1 and v2.0 \cite{squad1,squad2} are popular machine reading comprehension benchmarks. Their passages come from approximately 500 Wikipedia articles and the questions and answers are obtained by crowdsourcing. The SQuAD v2.0 dataset includes unanswerable questions about the same paragraphs.

\noindent $\bullet$ \textbf{BC2GM}. The Gene Mention Task at the Biocreative II workshop \cite{smith2008bc2gm} provides an annotated dataset for gene name entity recognition.

\noindent $\bullet$ \textbf{NCBI}. The NCBI disease corpus \cite{dogan2014ncbi} contains  annotations of disease mentions from a collection of PubMed abstracts.

\noindent $\bullet$ \textbf{JNLPBA}. JNLBA is a biomedical entity recognition shared task  \cite{collier-kim-2004-JNLPBA}. It is one of the largest datasets covering a large fraction of major taxonomies in molecular biology.

\end{document}